\documentclass[%
 aip,
cp,  
 amsmath,amssymb,
 reprint,%
]{revtex4-2}

\usepackage{graphicx}
\usepackage{dcolumn}
\usepackage{bm}

\usepackage[utf8]{inputenc}
\usepackage[T1]{fontenc}
\usepackage{mathptmx} 
\usepackage{multirow}
\usepackage{makecell}
\usepackage{multirow}

\begin{document}

\title{Knowledge Distillation Circumvents Nonlinearity for  Optical Convolutional Neural Networks}

\author{Jinlin Xiang} 
 \email{jinlinx@uw.edu}
 \affiliation{
    Department of Mechanical Engineering, University of Washington, USA}

\author{Shane Colburn}
 \email{scolbur2@uw.edu}
 \affiliation{
    Department of  Electrical \& Computer Engineering, University of Washington, USA}
 
\author{Arka Majumdar}
 \email[Corresponding author: ]{arka@uw.edu}
 \affiliation{
    Department of  Electrical \& Computer Engineering, University of Washington, USA}
\affiliation{
    Department of  Physics, University of Washington, USA}

\author{Eli Shlizerman}
 \email[Corresponding author: ]{shlizee@uw.edu}
 \affiliation{
    Department of  Electrical \& Computer Engineering, University of Washington, USA}
\affiliation{
    Department of  Applied Mathematics, University of Washington, USA}

\date{\today} 

\begin{abstract}
In recent years, Convolutional Neural Networks (CNNs) have enabled ubiquitous image processing applications. As such, CNNs require fast runtime (forward propagation) to process high-resolution visual streams in real time. This is still a challenging task even with state-of-the-art graphics and tensor processing units. The bottleneck in computational efficiency primarily occurs in the convolutional layers. Performing operations in the Fourier domain is a promising way to accelerate forward propagation since it transforms convolutions into elementwise multiplications, which are considerably faster to compute for large kernels. Furthermore, such computation could be implemented using an optical 4f system with orders of magnitude faster operation. However, a major challenge in using this spectral approach, as well as in an optical implementation of CNNs, is the inclusion of a nonlinearity between each convolutional layer, without which CNN performance drops dramatically. Here, we propose a Spectral CNN Linear Counterpart (SCLC) network architecture and develop a Knowledge Distillation (KD) approach to circumvent the need for a nonlinearity and successfully train such networks. While the KD approach is known in machine learning as an effective process for network pruning, we adapt the approach to transfer the knowledge from a nonlinear network (\textit{teacher}) to a linear counterpart (\textit{student}). We show that the KD approach can achieve performance that easily surpasses the standard linear version of a CNN and could approach the performance of the nonlinear network. Our simulations show that the possibility of increasing the resolution of the input image allows our proposed 4f optical linear network to perform more efficiently than a nonlinear network with the same accuracy on two fundamental image processing tasks: (i) object classification and (ii) semantic segmentation.
\end{abstract}

\maketitle

\section{INTRODUCTION}

Convolutional Neural Network (CNN) architectures are well known for their ability to compute visual  tasks~\cite{kakkava2018image,he2016deep,szegedy2015going,goodfellow2016deep,simonyan2014very}. Many of these tasks require fast processing of real-time inputs. In autonomous navigation, for example, a network must be capable of identify obstacles with different textures and lighting conditions in real time. While CNNs are instrumental in providing high accuracy for such tasks, the time that it takes for the input to propagate through the trained network (forward propagation time), is large and precludes real-time operation. The main reason for such inefficiency is the computational complexity of CNNs, which is $\mathcal{O}(HWk^2)$, where $H (W)$ is the height (width) of an image frame and $k\times k$ is the size of the convolutional kernel. The challenge of enhancing the computational performance has driven significant development of hardware that is dedicated to computing convolutions, with graphics processing units (GPUs) and tensor processing units (TPUs) that deliver an order of magnitude acceleration. Even with such dedicated hardware, however, effective computation times are still sub-optimal and become too large for many applications, especially when high-resolution inputs are processed. For example, when applied to ResNet-18 for autonomous navigation, the Jetson Nano, NVIDIA hardware development kit, achieves at most 5 frames per second (fps) rate for images of dimensions $1000\times1000$ \cite{mittal2019survey}. 

Since convolution is the most time-consuming operation, a possible gain in computational efficiency can be achieved by implementing the CNN in the Fourier domain (spectral domain)~\cite{mathieu2013fast,vasilache2014fast}. The Fast Fourier Transform (FFT) operation has the complexity of $\mathcal{O}(HWlog(HW)$ for the images and the kernels. In the Fourier domain, the convolution is transformed into an elementwise product with only $\mathcal{O}(HW)$ operations. While promising, in practice, the approach does not boost the forward propagation time. Due to nonlinearities that follow most of the layers in a CNN, the spectral approach ends up including a large number of costly forward and inverse Fourier transforms between successive layers. Optimizations of network architectures to be compatible with spectral operations were proposed, such as FCNN and Clebsch–Gordan Nets \cite{kondor2018clebsch,pratt2017fcnn}. These architectures, however, appear to suffer from a reduction in classification accuracy for complex visual tasks, for example, FCNN reaches less than half of state-of-the-art accuracy on CIFAR-10. Another branch of research developed in parallel is the introduction of spectral pooling layers for the purpose of adapting the spatially applied Max-Pooling operation to the spectral domain~\cite{rippel2015spectral,guan2019specnet,li2018frequency}. Unfortunately, these networks still entail nonlinear activations that require conversions between the spatial and spectral domains. Removing these nonlinear functions dramatically reduces the computational complexity of spectral CNNs but this often comes at the cost reduced performance. 

Another promising approach for accelerating computation is the development of optical neural networks (ONNs) that could replace electronic hardware~\cite{wetzstein2020inference}. ONNs utilize the inherent parallelism of light, which enables passive manipulation of massive amounts of data at the speed of light~\cite{cutrona1960optical,bueno2018reinforcement,lin2018all,colburn2019optical}. ONNs offer computation times that are nearly instantaneous in comparison to those provided by the best electronic hardware available. Specifically, a lens can perform a Fourier transform with $\mathcal{O}(1)$ complexity~\cite{goodman2005introduction}. In recent years, this promise has resulted in various ONNs for MNIST classification based on a sequence of diffractive masks in the Terahertz regime~\cite{lin2018all}, vector-matrix multiplication using integrated photonics~\cite{feldmann2021parallel,shen2017deep}, and hybrid optical-electronic networks leveraging the inherent Fourier transform property of lenses exploited in a $4f$ architecture~\cite{colburn2019optical}. While impressive in their own right, the demonstrated ONNs to date have been limited in terms of the complexity of the scenes on which they operate, which have mostly been limited to low-resolution images with simple features like MNIST digits. When applied to more complex scenes (e.g., CIFAR-10), these implementations exhibit low classification accuracy.

A major contributing factor to these limitations is that these networks lack the optical implementation of a nonlinear activation and were mostly constrained to linear operations in the optical domain~\cite{colburn2019optical}. Achieving an optical nonlinearity requires very large optical power, a high-quality factor resonator, exotic materials, or a combination thereof. While some nonlinear functions are readily available, such as the square operation imparted by a detector, it is unclear how effective it is for achieving comparable performance to that of the ReLU nonlinearity, which is the most common and effective nonlinear activation for CNNs~\cite{englund2012ultrafast,ryou2021free}. If the nonlinearity is implemented electronically as a subsequent layer, the electronic signal needs to be converted back into the optical domain to enable additional optical processing. Such repeated signal transductions significantly increase the power consumption and latency, thus obviating the benefits of an ONN versus a more traditional electronic implementation~\cite{colburn2019optical}.

To compensate for the lack of nonlinearity in an ONN, as well as to find an optimal level of performance for a fully linear spectral network, we develop a Knowledge Distillation (KD) training methodology to transfer the information from a nonlinear network (teacher) to a SCLC (student). Originally, KD training was introduced for pruning of networks, i.e., knowledge from a large teacher network is transferred to a less complex student model~\cite{hinton2015distilling,mishra2017apprentice}. Trained with the KD approach, the student network typically converges faster and obtains better performance than it would achieve without the KD training. A common example for successful KD training is object classification in images. In this problem, the teacher classifies images and provides "soft labels" to the student during training along with the actual labels. A Kullback-Leibler (KL) divergence loss between the soft labels and the student model predictions is then optimized to take into account the teacher's predictions~\cite{phuong2019towards}. KD training is a generic approach and was applied to a variety of problems such as semantic segmentation in which soft labels are used for classification of each pixel~\cite{li2017mimicking,liu2019structured}.

In this paper, we adapted the KD training framework to circumvent the need for a nonlinearity by "distilling the nonlinearity" from the nonlinear, teacher CNN, to the linear counterpart SCLC. We find that for boosting student accuracy, the teacher and the student networks are required to be as architecturally similar as possible.The KD approach allows the student network to achieve state-of-art performance, exceeding that of previous training methods or networks in the spectral domain, and is also easily amenable to optical implementation because light propagation is naturally described in Fourier space. The adapted KD training enables us to design a hybrid optical-electronic architecture for the SCLC network, where the optics serve as a linear frontend processing unit connected to an electronic backend that typically includes the last layer that corresponds to the task. We train this student network with KD training and demonstrate the performance of such a network on two common problems in which CNNs are leading computational methods: object classification and object segmentation . We show that the KD-trained SCLC can achieve performance easily surpassing that of a linear network trained with a standard training approach and nearing the performance of the nonlinear network.

\section{METHODS}
\subsection{Knowledge Distillation (KD)}
\begin{figure}[t]
\centering\includegraphics[width=1\textwidth]{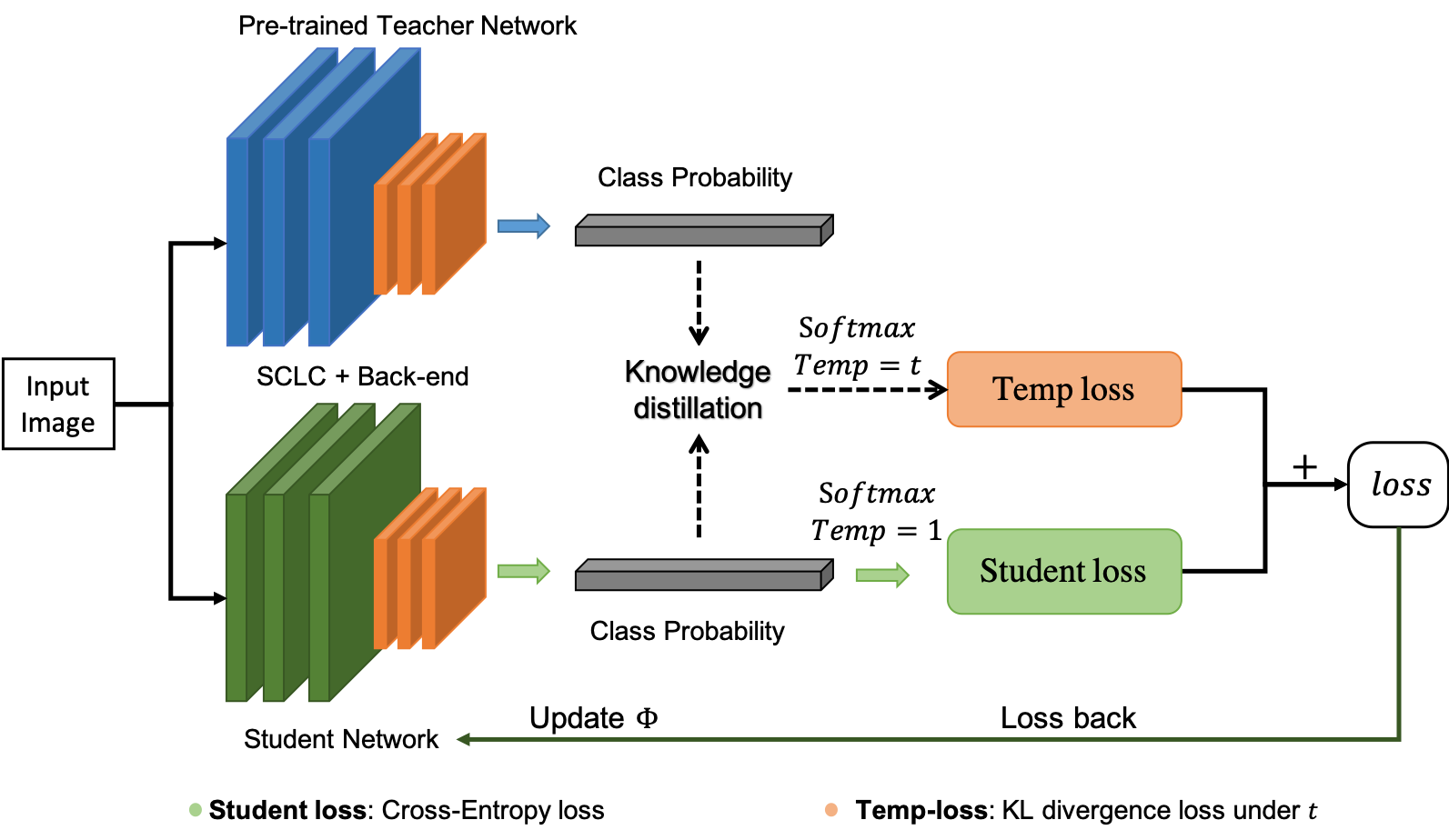}
\caption{Illustration of KD training. Student network (green-bottom) parameters $\Phi$ are updated according to an interpolation of two losses: the Student loss (Crossentropy loss between data and the student model output) and Temp loss (Cross-entropy loss or KL divergence) between the teacher network (blue-top) and student class distribution with a temperature parameter $T$. }
\label{KD}
\end{figure}

Knowledge Distillation is a machine learning method introduced for compression of neural networks~\cite{hinton2015distilling}. In particular, it defines the transfer of knowledge from a large neural network model, called the \textit{teacher}, to a small model, called the \textit{student}. We show the schematic flow of KD in Fig.~\ref{KD}. KD assumes that the teacher model is already trained and performs the given task with high accuracy. Then the student model is trained with both the conventional approach of minimization of the loss between the model prediction and the training data (\textit{Student loss}) and, in addition, compares the outcome of the student model with the teacher model through a temperature loss (\textit{Temp loss}) using the KL divergence as the loss~\cite{goodfellow2016deep}. The optimization of both losses is performed through the back-propagation, which adjusts the parameters of the student model using Stochastic Gradient Descent~\cite{bottou2012stochastic}, or ADAM~\cite{kingma2014adam} optimization. 

It was shown that student models trained with a KD approach can achieve significantly higher accuracy than the same model trained on the data alone without a teacher network~\cite{phuong2019towards}. The advantage of KD stems from the proposal to compute the temperature loss which treats the prediction of the teacher model as `soft labels' that inform the student model. In particular, conventional learning considers exclusively the labels in the training data as `hard labels' and for tasks such as classification computes the probabilities distribution vector $p^{hl}$, where each element $p^{hl}_i$ in the vector corresponds to the probability of the current input belonging to the class $i$. The softmax function is used to compute probabilities 
\begin{equation}
p^{hl}_i = exp(z_i)/ \sum_j exp(z_i).
\end{equation}
where $z_i$ is the student logits after the last fully connected layer. Training with hard labels is a sensitive process, especially for compact networks such as the student model. This typically results in inefficient networks and convergence to poor local optima. To overcome this limitation, KD proposes to add soft labels generated by the teacher model. These labels are 
the probabilities that the teacher model generates. In particular, for each input, at the same time of computing $p^{hl}$ with the student model, KD computes the soft probabilities vector, $p^{sl}$, with the teacher model according to
\begin{equation}
p^{sl}_i = exp(y_i/T)/ \sum_j exp(y_i/T).
\end{equation}
where, $y_i$ are the logits of the teacher after the last fully connected layer. Through $p^{sl}$ the teacher model contributes the probability estimates to the student model, knowledge that is unavailable from the data alone and is helpful for the student as providing extra information of the similarities between classes. The similarities are important since these indicate the effective knowledge of the teacher model and subsequently assists in achieving a similar knowledge and performance in the student model. The two probability distributions $p^{sl}$ and $p^{hl}$ are taken into account (as an interpolation) to compute the overall loss used in training of the student model.

The probabilities contributed by the teacher are defined as soft since the softmax function has a softening parameter, $T$, named as the distillation temperature. The success of KD depends on the choice of $T$. In a well-trained teacher model, the correct class has a much higher probability than other classes. For a low value of $T$, the probability of the correct class will approach $1$ while probabilities of other classes will be negligible and will not influence the training, due to $p^{sl}$ being similar to $p^{hl}$. On the other hand, when $T$ is too high, $p^{sl}_i$  will approach a uniformly distributed vector and will lose the distinction between the correct and incorrect classes. It is therefore important to chose $T$ in between these two extreme cases such that $p^{sl}$ would pass the similarities detected by the teacher to the student. 

In practice, the distillation of knowledge from the teacher to the student is particularly effective when the differences in the overall architecture between the two networks are minimal. In network compression, it is typically the case that the architectural blocks are kept the same and the only change that is implemented is the reduction of the number of neurons in each block. This leads us to the proposition of the implementation of KD between nonlinear and linear CNNs, where besides exclusion of nonlinear components, the linear network will be kept as similar as possible to the nonlinear one.

\subsection{The Spectral CNN Linear Counterpart (SCLC)}

\begin{figure}[t]
\centering\includegraphics[width=0.8\textwidth]{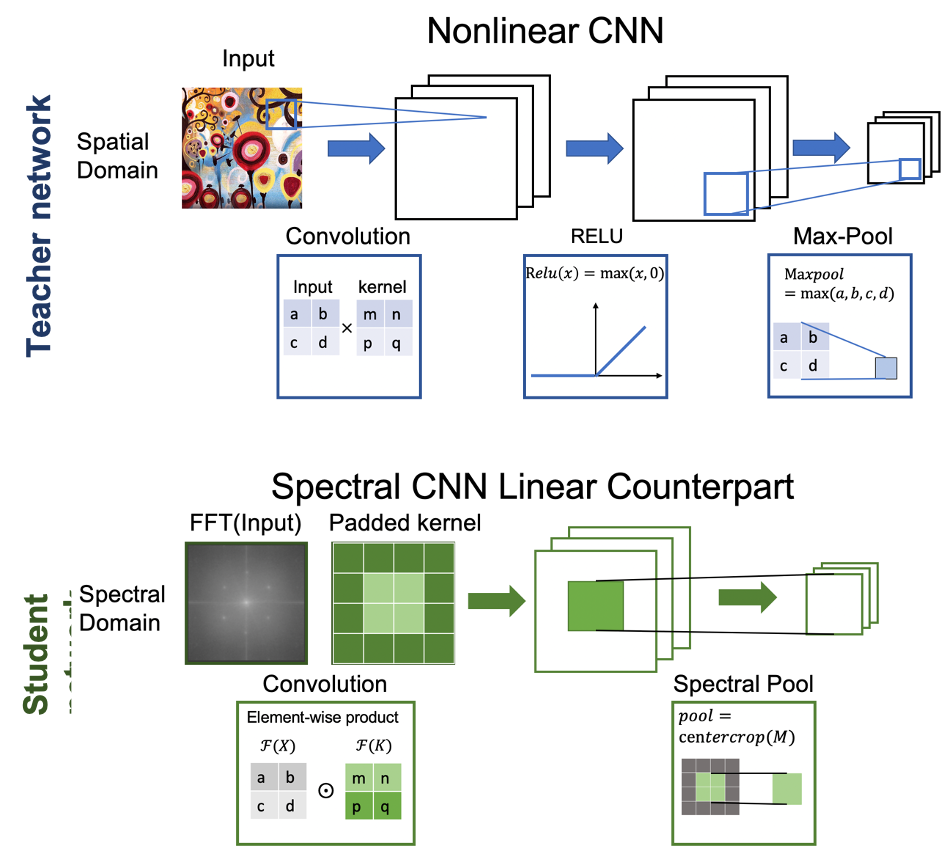}
\caption{Nonlinear CNN (top) and the proposed substitute, Spectral CNN Linear Counterpart (SCLC) (bottom). Top: An exemplar nonlinear CNN with layers that include operations of Convolution, Nonlinear RELU activation, and Max-Pool. 
Bottom: SCLC of the CNN shown in top row. The Convolution corresponds to the elementwise product in the spectral domain. The RELU operation is excluded. The Max-Pool layer is represented by a center-crop operation in the spectral domain.}
\label{SCLC}
\end{figure}

We propose to construct a Spectral CNN Linear Counterpart (SCLC), the "Student Network", to a spectral nonlinear CNN, the "Teacher Network", based on an optical $4f$ architecture. The "Teacher Network"  takes the shape of a common CNN designed for generic tasks, for example, image classification or object segmentation. In the case of classification, we consider CNNs that are structured with multiple layers of repeating operations of Convolution, Nonlinearity (ReLU), and Max-Pool (selecting the maximal value), as demonstrated in Fig.~\ref{SCLC}. In the case of object segmentation, we consider a CNN of a ``U" shape~\cite{ronneberger2015u}, where the input passes through similar operations of Convolution, ReLU, and Max-Pool, and in addition, the output of each layer contracts the input dimension up to a ``bottle-neck" layer from which the representation is expanded with a set of inverse operations such as Up-Convolution, ReLU, and Max-Pool. To measure the performance of the proposed architectures, we concatenate them with a backend that corresponds to either a classification task (softmax fully connected backend layer) or a segmentation task (sequence of Up-Convolution backend layers).

To obtain a network model that is realizable with optical components, the network requires a conversion to the spectral domain, such that the input into the model is the Fourier transform of the input image and the operations such as convolution and pooling are implemented in the spectral domain~\cite{colburn2019optical}. Furthermore, the model should not include nonlinearities since computing those will require an inverse transform and transition from optical to electronic components. We describe the building blocks of such a proposed model below.
\subsubsection{Convolution in SCLC}
The convolution is performed as the elementwise product between the input images represented in the spectral domain and the kernels padded to the same dimensions as the inputs. The input into the convolution is a four-dimensional tensor (S: batch size, C: number of input channel, H: height, W: width), and the 2D convolution operation with a stride of 1 is calculated via
 \begin{equation}
 y(i,j) =x*k= \sum^{H-1}_{h=0} \sum^{W-1}_{w=0} x(h,w) \times k(i-h,j-w)
 \end{equation}
 where $x$ is the input image, $k$ is the kernel and $*$ indicates the Convolution operation. We implement the convolution in SCLC in the spectral domain by using an FFT and an elementwise product
 \begin{equation}
 Y=\mathcal{F}(x*k) = \mathcal{F}(x) \odot \mathcal{F}(k) \label{eq:specconv}
 \end{equation}
The function $\mathcal{F}$ denotes the FFT operation, and $\odot$ indicates the elementwise product, which requires the input and the kernels to have the same dimension $H\times W$. During training, the spectral convolution kernel is updated according to
\begin{equation}
\begin{aligned}
&\sigma_X = \nabla_X\mathcal{L}|_{X=X_0} = \sigma_Y \odot K_0 \\
&\delta_k = \nabla_K\mathcal{L}|_{K=_{\mathcal{F}(0)}} = \sigma_Y \odot X_0 \\
&k_1 = k_0 + \lambda[\mathcal{F}^{-1}(\delta_K)], 
\end{aligned}
\end{equation}
where $\sigma_X (\sigma_Y)$ is the error from the previous (next) layer; $X_0$ and $K_0$ are, respectively, the input and kernel in the forward propagation; $\nabla_K (\nabla_X)$ is the gradient operator w.r.t. the kernel (input); and $\mathcal{L}$ is the loss function. The updates to the elements in the kernel are computed by applying an inverse FFT (i-FFT) and multiplying by the learning rate $\lambda$. 

Once the network has been trained, the implementation with spectral convolution components has significant benefits for inference latency for a given input (forward propagation). Indeed, assuming an input image size of $(H,W)$ and square kernel size of $(k,k)$, the complexity of one spatial convolution in the image domain is $\mathcal{O}(HWk^2)$. On the other hand, the spectral convolution composed of elementwise products is of complexity $\mathcal{O}(HW)$. In the optical setup, elementwise operations can be performed in parallel and thus the runtime, being independent of $H$ and $W$, is further reduced to $\mathcal{O}(1)$. In addition, the computational complexity of an FFT is $\mathcal{O}(HWlog(HW))$, which brings the overall complexity for the spectral convolution to be $\mathcal{O}(HWlog(HW))$~\cite{lee2018single,pratt2017fcnn,kappeler2017ptychnet}. As the proposed network is in the spectral domain, the FFT and i-FFT transforms are needed to be applied at the beginning and at the end of the network for images only, such that the complexity of the intermediate layers will only apply FFT to kennels, which greatly reduces the transformation times. Additionally, the main acceleration occurs in the optical implementation since the spectral transforms can be achieved through phase transforming components which operate at the speed of light. The combination of convolution and transform in optical domain would correspond to instantaneous running time of $\mathcal{O}(1)$, when compare to electronic running time for all layers of the network.

\subsubsection{Pooling in SCLC}
Implementation of pooling in the spectral domain needs to take into consideration both mimicking the effect of the standard Max-Pool used in CNNs and to be a practical implementation with optical components. In a nonlinear CNN, common pooling operations are Average-Pooling or Max-Pooling, which select the average of the elements or the maximum element, respectively, from the region of the feature map covered by the pooling filter shown in Fig.~\ref{SCLC}. The purpose of the pooling layers is to reduce the dimensions of the feature maps and to find the representative elements to be transferred forward from the feature map. We propose to substitute the Max-Pool with a different pooling function, called ``Spectral pool",  which will enable similar functionality in the spectral domain. In particular, we propose to replace the pooling layer with a linear, low-pass filter in the spectral domain. The forward and backward propagation in the layer are defined as 
\begin{equation}
    y=\mathcal{F}^{-1}(CROP(\mathcal{F}(x),(H',W')))
\end{equation}
\begin{equation}
    z=\mathcal{F}^{-1}(PAD(\mathcal{F}(y/x^*),(H,W)))
\end{equation}
where the $CROP$ operation in forward propagating keeps the center part of the spectral feature map and reduce the total size from $(H,W)$ to $(H',W')$. The $PAD$ operation in backward propagation matches the dimension of gradients outputs from $(H',W')$ to $(H,W)$ by padding zeros instead of cropped elements. The $\mathcal{F}$ and $\mathcal{F}^{-1}$are applied if the networks are in spatial domain. Those operations are similar to max pooling since the idea of Max-Pool is to find the representative element in each kernel, while it goes over the feature map and effectively selects the most descriptive features. Spectral-Pool drops the high frequencies in the Fourier domain such that it achieves a less noisy output and captures the dominant features, similar to the effect of the conventional Max-Pool operation. It was shown that training with spectral pooling has better convergence properties compared to Max-Pooling~\cite{rippel2015spectral}. Furthermore, a spectral pooling implementation is of lower computational complexity and is furthermore inherently amenable to optical implementation as light is well described in the Fourier domain $\mathcal{O}(1)$ complexity, see Table~\ref{tab:table1}.
\begin{table}
\begin{ruledtabular}
\caption{\label{tab:table1} Comparison of forward propagation computational complexity between Nonlinear CNN, SCLC implemented on electronic hardware, SCLC implemented on optical hardware}

\begin{tabular}{llll}
Structure                            & Nonlinear CNN                         & SCLC (Computational)                   & OSCLC (Optical)    \\  \hline
\multirow{2}{*}{Convolution layers } & \multirow{2}{*}{$\mathcal{O}(HWk^2)$} & Spatial: $\mathcal{O}(HW \log (HW))$   & \multirow{2}{*}{$\mathcal{O}(1)$}  \\
                                     &                                       &  Spectral: $\mathcal{O}(HW \log (HW))$ &                    \\  \hline
\multirow{2}{*}{Pooling layers }     & \multirow{2}{*}{$\mathcal{O}(HWk^2)$} &  Spatial: $\mathcal{O}(k^2 \log (k^2))$& \multirow{2}{*}{$\mathcal{O}(1)$}  \\
                                     &                                       & Spectral: $\mathcal{O}(1)$             &                   
\end{tabular}
\end{ruledtabular}
\end{table}

Typically, an activation function, such as ReLU, $\tanh$, or $\sigma$ is applied to extract the features after convolution to avoid extremities in neural units values and maintain network stability. These nonlinearities, however, are nearly impossible to achieve in a practical optical implementation. Thus, the SCLC skips the nonlinear activation function to make the network structure amenable to optical implementation. The lack of nonlinearity will be circumvented by the application of KD to maximize the performance achieved using linear operations only.

\subsection{Knowledge Distillation in SCLC}
KD training requires a teacher and student model both performing the same task and exhibiting similarity in their architectures. In previous applications, KD training was applied to model pruning, where the teacher and the student models have exactly the same structure with the student having a fraction of the units of the teacher. Here, we extend KD application and consider the teacher and the student models with a similar number of units; however, the student is an adapted version of the teacher operations lacking the nonlinear activation and has a revised pooling operator. In particular, the teacher model is a pre-trained nonlinear CNN, which is a state-of-the-art system for that particular task. The student network is the SCLC that corresponds to that system with the architectural changes described in the previous sections. The objective of KD training is to optimize the total loss subject to updating the weights, $\Phi$, of the SCLC student model only. The total loss is a linear combination of the Temperature loss and the Student loss. The weights are updated according to implementation of back-propagation that optimizes the total loss.

We select the Temperature loss to be the KL function between the soft labels ($p^{sl,t}$) from the pre-trained nonlinear CNN model (\textit{teacher}) and predictions ($p^{sl,s}$) from SCLC (\textit{student}) both distilled with the same temperature $T$. We chose KL loss over cross entropy because it includes an extra penalty on the direction of the loss, which facilitates convergence. The student loss is the standard cross-entropy loss between the data labels and SCLC probabilities ($p^{hl}$). The total loss is then calculated as a weighted summation of the two losses
\begin{equation}
Loss(x,\Phi)=\alpha \mathcal{L}_C(y,p^{hl})+(1-\alpha)\mathcal{L}_K((p^{sl,t};T=\tau),(p^{sl,s};T= \tau)),
\end{equation}
where $x$ corresponds to the input, $y$ is the training data, $\Phi$ are student model weights, $\mathcal{L}_C$ is the cross-entropy loss function, $\mathcal{L}_K$ is the KL divergence loss function, $p^{hl}$ corresponds to the student model hard predictions, $p^{sl,s}$ corresponds to the student predictions under given teacher model probabilities $p^{sl,t}$, and $\alpha$ is the weighting parameter.

\subsection{Optical Implementation of SCLC (OSCLC)}
In this section, we explore how the proposed SCLC network can be implemented using free-space optics, i.e. Optical SCLC (OSCLC). The choice of free-space optics is motivated by the large number of information channels available to us in such an implementation~\cite{ryou2021free}. There are three components of the OSCLC that need to be considered for optical implementation: convolutional layers, spectral pooling, and summation of different channels.

Each convolutional layer can be implemented by a $4f$ correlator architecture to further accelerate the forward propagation speed \cite{cutrona1960optical,psaltis1995holography,lu1989two,psaltis1988adaptive}. A typical $4f$ correlator architecture comprises two lenses of equal focal length spaced apart at $2f$ distance and with input and output planes located in the front and the back focal planes of the first and second lenses respectively. The first lens produces a Fourier transform of the input scene at the focal plane. A mask placed in that focal plane provides the point-wise multiplication implementing the convolution and the second lens performs the inverse Fourier transform. Therefore, a $4f$ correlator architecture is able to work as an equivalent architecture for a single channel of a linear spectral CNN counterpart. The computational complexity can decrease to $\mathcal{O}(1)$ for any operations in spectral domain and will not be exponential growth with the input image resolution/ pixels. The spectral pooling in the OSCLC can also be implemented by a low-pass filter, where we block high frequency components in the Fourier plane. 

The summation of different channels is a basic principle in deep neural networks and is widely used in object classification and segmentation. If there are no summations after elementwise products, the channels will grow exponentially, requiring a massive number of kernels after only a few convolutional layers, which makes it impossible for real optical implementations. Although fewer kernels  in the convolution layers can alleviate this challenge, the overall accuracy, especially in complex scenarios, would suffer. Such a summation can be implemented using various techniques used for coherent beam combining~\cite{gerke2010aperiodic,zhan2017metasurface}. We note that, while free-space optics tends to be bulky and prone to misalignment, recent demonstrations of meta-optics and volume optics~\cite{prossotowicz2020coherent,fsaifes2020coherent} exhibit complicated free-space optics in a compact form factor, possibly in a monolithic fashion mitigating any misalignment.

\section{COMPUTATIONAL EXPERIMENTS}

We evaluate our proposed SCLC on two distinct tasks involving image inputs: (i) object classification and (ii) object segmentation. Object classification task associates each image with a class which corresponds to the object photographed in the image, for example, the task could be to classify whether the image depicts a cat, a dog, or a car. Object segmentation is a pixel-level classification task. It aims to determine for each pixel in a given image to which class it belongs. Effectively, object segmentation partitions the input image into regions that correspond to classes. For example, in the case of a photograph of a person, regions would correspond to the background, face, hair, etc. Our evaluation consists of determining the accuracy and the forward propagation runtime that the SCLC and OSCLC achieve compared to a standard CNN. 

We test the performance metrics on a variety of common benchmarks for each task. For object classification, we perform experiments on (i) Kaggle Cats and Dogs Challenge~\cite{catdog}, (ii) Cifar-10~\cite{krizhevsky2009learning}, and (iii) HIGH-10 (a subset of ImageNet)~\cite{krizhevsky2012imagenet}. For object segmentation, we perform experiments on (i) Kaggle's Carvana Image Masking Challenge (Cars Segmentation)~\cite{carseg}, (ii) Face Recognition \cite{shen2016deep} and (iii) VOC2012~\cite{pascal-voc-2012}. Training of all networks is performed on a Tesla P100 GPU with the Google Colab platform. For all networks, the initial learning rate for student networks is 0.0001 with momentum 0.9 and weight decay 0.0005. We set the batch size according to the available memory: 16 for the classification task and 1 for the segmentation task. We use the cross-entropy loss to calculate the student loss, and KL divergence to calculate the Temp loss.

\subsection{Spectral Counterpart Layer Efficiency}
\begin{figure}[htbp]
\centering\includegraphics[width=1\textwidth]{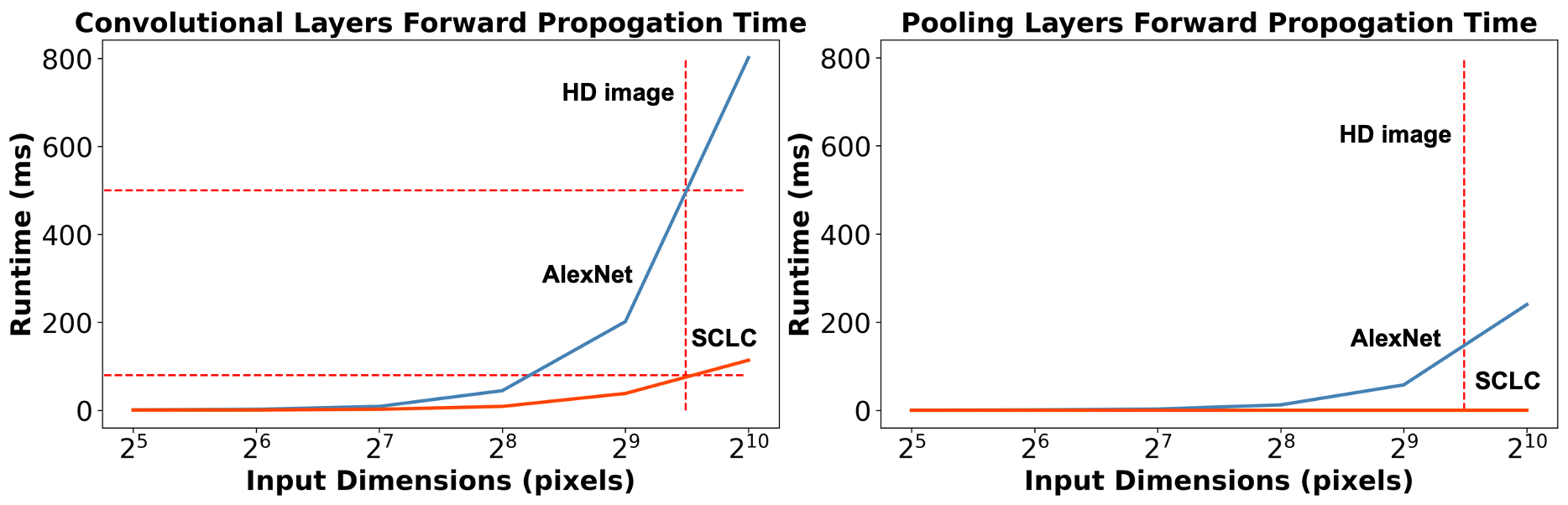}
\caption{Comparison of forward propagation time between AlexNet (blue line) and its SCLC (red line). Left: Propagation time through a single convolutional layer for variable image dimensions; Right: Propagation time through a single pooling layer for variable image dimensions. The dashed lines mark the resolution of an HD image and its corresponding runtime in AlexNet and SCLC.}
\label{dvst}
\end{figure}

We first investigate the runtime of forward propagation when variable input dimensions ($32\times32$ to $1024\times1024$) are considered for AlexNet (nonlinear CNN) vs. SCLC, and show our results in Fig.~\ref{dvst}. We evaluate forward propagation time for convolutional layers and pooling layers separately. Forward propagation time of convolutional layers in AlexNet grows exponentially with input size. The runtime of SCLC grows exponentially as well but with a significantly slower rate. For example, For High Definition (HD) input size image, the runtime of SCLC convolution corresponds to 100ms vs. 500ms for AlexNet convolution. For pooling layers, AlexNet forward propagation runtime grows exponentially with input size with a smaller rate than the convolution. For SCLC, due to the pooling being spectral, the runtime is constant and input independent. 

\subsection{Object Classification Task}
We evaluate our proposed SCLC on different classification datasets to estimate the accuracy that SCLC can achieve when trained with KD. We compare the accuracy to that of the teacher network, AlexNet. The first dataset that we consider is the Kaggle Cats and Dogs Challenge~\cite{catdog} which consists of $125000$ images of dimensions $96 \times 96$ associated with two classes: a cat or a dog. Additional dataset that we consider is Cifar-10~\cite{krizhevsky2009learning} which consists of $60000$ images of dimension of $32 \times 32$ including 10 classes. The third dataset that we consider is High-10 which is a subset of ImageNet~\cite{krizhevsky2012imagenet} and consists of approximately $10000$ annotated images, equally distributed in $10$ classes with images of resolution $500 \times 300$. During training of AlexNet all images are resized or cropped to dimensions of $224 \times 224$ to match AlexNet's input size. To evaluate the performance of SCLC in classification we add a single fully connected backend layer to it. The backend layer is intended to be implemented with electronics in OSCLC since includes nonlinearity.
We employ the KD approach to train SCLC with AlexNet being the teacher network and when the training converges we test SCLC variants against AlexNet. 

In particular, we compare the SCLC trained with and without the KD approach or a variation of AlexNet with square nonlinearity that could be realized optically with 4f components. We show the results of the comparison for the three benchmarks in Table~\ref{tab:table3}. We observe that for all benchmarks, KD training significantly enhances the accuracy of the classification achieved by the SCLC (by $ 12.5\%$ on average). Indeed, KD contribution appears to be essential in generating an SCLC network with robust accuracy. On Kaggle's Cats and Dogs classification SCLC achieves 90.60\% ($6\%$ below the accuracy of AlexNet), on Cifar-10 classification it achieves 80.80\% ($5\%$ below the accuracy of AlexNet) and on HIGH-10 classification it achieves 81.4\% ($\approx 12\%$ below the accuracy of AlexNet). These results are encouraging since SCLC trained with standard training has a much bigger gap of $16\%, 20\%, 23\%$ between its accuracy and the accuracy of AlexNet. The KD approach appears to close this gap by more than half for all benchmarks. Furthermore, when the RELU nonlinearity is modified to square nonlinearity (SQ-AlexNet) both KD and standard training do not result with sufficiently accurate network. This is because the square nonlinearity will magnify the parameters of the model. This observation indicates that the KD approach is effective in regimes where the network architecture of the student is kept as close to the teacher as possible.

\begin{figure}[htpb]
\centering\includegraphics[width=1\textwidth]{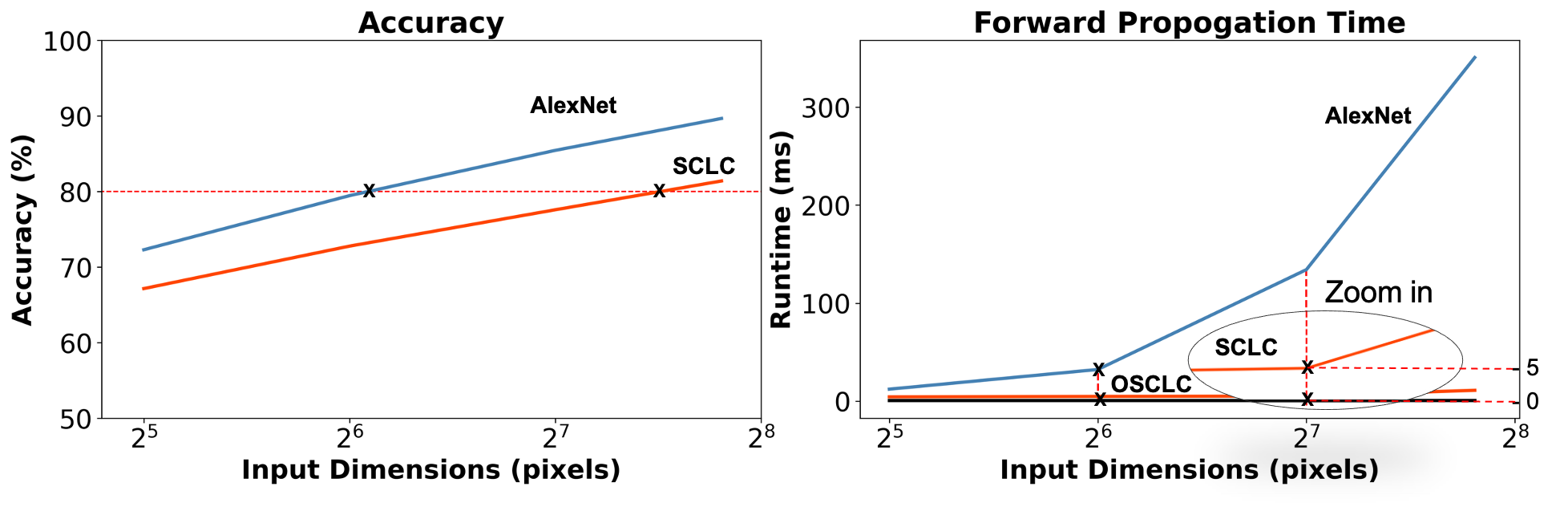}
\caption{Accuracy and Forward propagation time for classification task on High-10 dataset. Left: AlexNet and SCLC accuracy for varying input resolution; Right: Forward propagation runtime of AlexNet, SCLC, and OSCLC for varying input resolution. 
}
\label{clas}
\end{figure}

Notably, to match AlexNet input dimensions, all experiments were implemented with the same input image resolution of $224\times224$. We therefore explore with HIGH-10 dataset (that includes higher resolution images) how accuracy varies if the resolution of the input is increased. For each input image resolution down sampling to $224\time244$, the teacher (AlexNet) is first trained and then the student (SCLC/OSCLC) is trained under the supervision of the teacher model. We show in Fig.~\ref{clas} that increasing the resolution of the input increases the accuracy of both the teacher and student models (with a linear rate). Consequently, SCLC trained with higher resolution inputs can surpass the accuracy of the teacher network trained with lower resolution dataset, e.g., SCLC with input of $2^{7} \times 2^{7}$ dimensions performs similarly ($\approx 80\%$) to the teacher with input of $2^{6} \times 2^{6}$ dimensions. While the accuracy of the two is similar, the runtime of SCLC is expected to be more favorable since teacher's runtime grows exponentially. Indeed, we show that for $80\%$ accuracy both SCLC and OSCLC are at least 5 times faster than ($\approx 5ms$ and $\approx 1ms$) than AlexNet ($\approx 25ms$).

We further demonstrate the efficient runtime of SCLC compared to AlexNet in Table~\ref{tab:table3} columns 3,4 for input size of $224 \times 224$. While AlexNet runtime for a single image is $350ms$, SCLC runtime drops by an order of magnitude to $11ms$ for a single image. Simulated runtime of OSCLC drops by another order of magnitude to $0.6ms$. The estimation of OSCLC forward propagation runtime includes three parts: 1) running time of the optical structure, 2) transduction time between optics and electronics, and 3) running time of the electronic backend. Since light propagation is very short, $\approx ps$, the main contribution comes from the signal transduction ($\approx 0.32 ms$ for 100kb images via USB 3.0 protocol at a rate of 2500Mbit/s) and the backend propagation time ($\approx 0.28ms$ for a single image on a GPU (Tesla P100).

\begin{table}
\begin{ruledtabular}
\caption{\label{tab:table3}Accuracy and Estimate the efficiency in forward propagation in classification}
\resizebox{\columnwidth}{!}{%
\begin{tabular}{cccccc}
Dataset                           & Model(input:$224 \times 224$)  & Accuracy                         &    Runtime (ms/img)  &  Optical Runtime (ms/img)\\ \hline
\multirow{5}{*}{Cats vs. Dogs}    & AlexNet                     & $96.10\%$                        & 350.66               & - \\
                                   & SCLC                        & $79.50\%$                        & 11.05                & 0.61 \\
                                  & SCLC + KD                   & $90.60\%$ (\textbf{$+11.10\%$})  & 11.05                & 0.61  \\
                                  & SQ-Nonlinear                & $51.70\%$                        & 12.04                & 0.61  \\
                                  & SQ-Nonlinear+ KD            & $51.72\%$                        & 12.04                & 0.61   \\\hline
 \multirow{3}{*}{Cifar-10}         & AlexNet                     & $85.09\%$                        & 350.66               & -   \\
                                   & SCLC                        & $65.45\%$                        & 11.05                & 0.61 \\
                                   & SCLC+KD                     & $80.80\%$ (\textbf{$+15.35\%$})  & 11.05                & 0.61 \\\hline
 \multirow{3}{*}{High-10}          & AlexNet                     & $93.95\%$                        & 350.66               & -     \\
                                   & SCLC                        & $70.12\%$                        & 11.05                & 0.61 \\ 
                                  & SCLC + KD                   & $81.40\%$ (\textbf{$+11.28\%$})  & 11.05                & 0.61 \\
\end{tabular}}
\end{ruledtabular}
\end{table}

To demonstrate how different components contribute to SCLC performance, we investigated ablations of the SCLC architecture (Table~\ref{tab4}). We first compared Spectral Pooling with Max Pooling in the SCLC. We find that Spectral Pooling corresponds to faster convergence and to 1.71\% higher accuracy than Max Pooling. In addition, we test the contribution of the frontend (SCLC), the backend and the KD training of them. When only the backend is considered, the accuracy drops to 41\%. Adding the frontend to it and training the model on the same dataset corresponds to an increase of $\approx 30 \%$  such that the accuracy becomes $70.12\%$. Further application of KD training of both the frontend and the backend boosts the accuracy and convergence, such that the accuracy is enhanced by another $\approx 10 \%$ to $81.4\%$.
\begin{table}
\begin{ruledtabular}
\caption{Ablation studies of pooling components, SCLC, backend and KD training}
\begin{tabular}{cc} 
Structure                    & Accuracy \\ \hline
Max Pooling         & 68.41$\%$  \\
Spectral pooling                         & 70.12$\%$(+1.71$\%$) \\ \hline
Backend only           & 41.40$\%$           \\
SCLC (Frontend) + Backend    & 70.12$\%$ (+28.72$\%$) \\
SCLC (Frontend) + Backend + KD & 81.40$\%$ (+11.28$\%$) \\ 
\end{tabular}
\end{ruledtabular}
\label{tab5}
\end{table}

\subsection{Object Segmentation Task}
\begin{figure}[htbp]
\centering\includegraphics[width=0.5\textwidth]{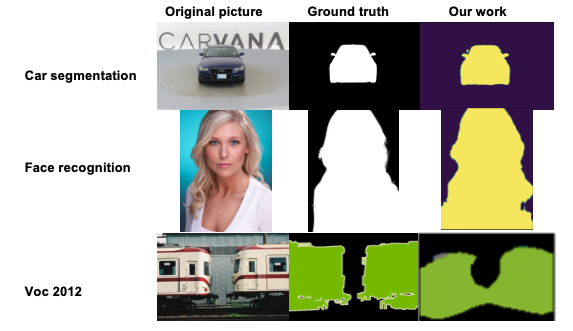}
\caption{Examples of object segmentation for the considered benchmarks of (i) Car Segmentation, (ii) Face Recognition, and (iii) VOC2012. Left to right: Input image from datasets, Ground truth, Our work (SCLC/ OSCLC).
}
\label{fig3}
\end{figure}

In addition to image classification, we explore the construction of an SCLC for object segmentation as shown in Fig.~\ref{fig3}. For such a task, a standard CNN is of a U shape with a sequence of convolutions and up convolutions. We choose the teacher network to be a  U-Net~\cite{ronneberger2015u}. KD training corresponds to pixel-level loss for both soft predictions and soft labels. The frontend for this task is the sequence of contracting convolutions while the backend corresponds to a sequence of up convolutions. 

The first benchmark that we consider is the Kaggle’s Carvana Image Masking Challenge that consists of 5088 cars with an original resolution of $1920\times 1280$. The dataset is randomly split into 4580 and 508 images for training and testing, respectively. Our second benchmark is the Face Recognition dataset, whichconsists of 2000 images with 1700 for training and 300 for testing. Our third benchmark is VOC2012, which consists of 2913 images of original resolution $500 \times 375$, which includes 20 classes and one background class. The images in all datasets are down-sampled to $960 \times 640$ or smaller due to limitations of GPU memory.

Our results are shown in Table~\ref{tab4}. For both Car and Face datasets we observe that the SCLC performs relatively well and obtains accuracy that falls from that of the teacher U-Net by only a few percent. We observe that consideration of higher resolution inputs corresponds to enhanced accuracy in classification experiments. For the VOC2012 dataset, which includes lower resolution images and has more segmentation classes, both the teacher and SCLC perform with rather low accuracy below $80\%$. This example demonstrates that in such problems, it is crucial to consider the fully available image resolution. We compare the computational efficiency of the SCLC against the teacher in terms of frames per second rate and find that the SCLC is approximately $2\times $ faster than U-Net. While such a speedup is more modest than in the image classification task, the rate is closer to a real-time operation rate (30 fps) or alternatively allows for consideration of larger input images that may correspond to enhanced segmentation with the same frame rate as that of the teacher. The main contribution to SCLC runtime is the electronic backend that consists of up-convolution layers, which have a computational complexity of $\mathcal{O}(HWk^2)$ (compared to image classification backend, which is a fully connected single layer). 

\begin{table}
\begin{ruledtabular}

\caption{\label{tab4}Accuracy and forward propagation rate (frames per second) for object segmentation task tested on three benchmarks.}
\begin{tabular}{cccc}

Dataset                           & Network          & Accuracy & Rate (higher is better) \\ \hline
{Car Segmentation} & U-Net           & 98.02\%  & 7.36 fps     \\
                                  &  OSCLC & 97.1\%   & 12.7 fps     \\ \hline
{Face Recognition} & U-Net           & 95.58\%  & 9.38fps      \\
                                  & OSCLC & 91.39\%  & 15.6fps      \\ \hline
{VOC2012}         & U-Net           & 75.51\%  & 7.36fps      \\
                                  & OSCLC & 62.21\%  & 12.7fps      \\ 
\end{tabular}
\end{ruledtabular}
\end{table}

\section{CONCLUSIONS}



Over the last decade, various CNNs have been developed and deployed as robust systems for ubiquitous visual tasks. While such incorporation in multiple applications is remarkable, scaling up CNNs for tasks that require high speed and energy efficiency entail significant challenges. This is due to the fact that the computational runtime of convolutional layers increases exponentially with input size. Such a constraint eventually leads to CNNs exceeding the limit of power consumption and as such they limit the accuracy and rate at which the system can perform. It is often the case that these limitations prohibit real-time deployment of CNNs.
Representation of CNNs in the spectral domain in conjunction with an optical platform that supports parallel, elementwise product computations has the potential to overcome the exponential increase in computation runtime. However, since transitions from optical to electronic systems are computationally ineffective, it is required to compute the additional, typically nonlinear, layers of CNNs with the same optical platform. Performing such operations with optical components is currently impractical. 
To that end, we propose a spectral linear version of CNN (SCLC) that is also optically realizable (OSCLC) with 4f components and investigate whether it can serve as an effective counterpart for CNN. While it is indeed possible to design such a system, we show that the accuracy significantly drops when nonlinear activations are not present.
To address the gap in accuracy between CNN and its SCLC, we propose a novel training approach based on Knowledge Distillation (KD). The approach assists in training the SCLC by involving the original CNN in the training as the teacher in addition to the standard optimization with the training data. 
We show that such training is promising in circumventing the need for a nonlinearity and represents a generic training procedure that could be applied to various CNNs and their respective SCLCs. In particular, we consider two visual tasks and six benchmarks and show that KD training applied to training a SCLC boosts its accuracy by more than half of the gap created by exclusion of nonlinearities.
We further show that increasing input dimensions has almost no impact on computational efficiency; however, increasing dimensions is able to further close the accuracy gap or even surpass the accuracy of CNN operating with low resolution inputs. 

While our focus here is on realizing an optically viable architecture for an SCLC, it is notable that our analysis and experiments indicate a computational benefit for an SCLC trained with KD even for standard electronic components such as a GPU. 
For the OSCLC, we find that keeping the backend layer, which is implemented electronically, as minimal as possible is critical to fully make use of the benefits of optical parallelism and its speed. Indeed, as we demonstrate in our simulations, in the image classification task, in which the backend consists of a single layer, optical forward propagation reduces the runtime for standard input image dimensions by two orders of magnitude. In the object segmentation task, the backend is more complex. It consists of up-convolutions and thus while $\times 2$ computational speedup is achieved, it is not as remarkable as the speedup in the case of the classification task. 
Furthermore, our investigation indicates that  key components of CNNs, such as convolution and pooling, can indeed be adapted to the optical domain and for such adaptation the training process needs to be an integrated training procedure that involves both the original CNN, its adapted counterpart, and the electronic backend. We show that KD training presents a plausible integrated training procedure for these purposes. 
Adaptation of additional components alongside further development of KD-based training may pave the way toward optical-electronic deployment of CNNs to provide a further leap in enhancement of CNN performance for complex and real-time tasks.

\section{Acknowledgement}
A.M. and S.C. are supported by Washington Research Foundation, UW Reality Lab, Facebook, Google, Amazon, and Futurewei. E.S. acknowledges partial support of Washington Research Foundation and the departments of Applied Mathematics and Electrical and Computer Engineering at the University of Washington.

\nocite{*}

\providecommand{\noopsort}[1]{}\providecommand{\singleletter}[1]{#1}%

\end{document}